\documentclass[10pt,twocolumn,letterpaper]{article} 

\usepackage{avss}
\usepackage{times}
\usepackage{epsfig}
\usepackage{graphicx}
\usepackage{amsmath}
\usepackage{amssymb}

\usepackage{array}
\usepackage{soul}
\usepackage{multirow}

\usepackage{algorithm} 
\usepackage{algpseudocode}

\usepackage{kotex}
\usepackage{xcolor}
\usepackage{pifont}

\usepackage{blindtext}
\usepackage{epstopdf}

\newcommand{\myfootnote}[1]{
    \renewcommand{\thefootnote}{}
    \footnotetext{\scriptsize#1}
    \renewcommand{\thefootnote}{\arabic{footnote}}
}

\newcolumntype{L}[1]{>{\raggedright\let\newline\\\arraybackslash\hspace{0pt}}m{#1}}
\newcolumntype{C}[1]{>{\centering\let\newline\\\arraybackslash\hspace{0pt}}m{#1}}
\newcolumntype{R}[1]{>{\raggedleft\let\newline\\\arraybackslash\hspace{0pt}}m{#1}}

\definecolor{mypurple}{rgb}{0.427, 0.137, 0.510}

\newcommand{\cmark}{\ding{51}}%

\usepackage[pagebackref=true,breaklinks=true,letterpaper=true,colorlinks,bookmarks=false]{hyperref}

\avssfinalcopy 

\def\avssPaperID{53} 

\ifavssfinal\fi
\begin{document}

\title{Position-aware Location Regression Network for Temporal Video Grounding}


\author{Sunoh Kim$^1$ \hspace{1.5cm} Kimin Yun$^2$ \hspace{1.5cm} Jin Young Choi$^1$\\
$^1$Department of ECE, ASRI, Seoul National University, South Korea\\
$^2$Electronics and Telecommunications Research Institute (ETRI), South Korea\\
{\tt\small \{suno8386, jychoi\}@snu.ac.kr, kimin.yun@etri.re.kr}
}

\maketitle
\thispagestyle{empty}

\begin{abstract}

The key to successful grounding for video surveillance is to understand a semantic phrase corresponding to important actors and objects.
Conventional methods ignore comprehensive contexts for the phrase or require heavy computation for multiple phrases.
To understand comprehensive contexts with only one semantic phrase,
we propose Position-aware Location Regression Network~(PLRN) which exploits position-aware features of a query and a video.
Specifically, PLRN first encodes both the video and query using positional information of words and video segments.
Then, a semantic phrase feature is extracted from an encoded query with attention.
The semantic phrase feature and encoded video are merged and made into a context-aware feature by reflecting local and global contexts.
Finally, PLRN predicts start, end, center, and width values of a grounding boundary.
Our experiments show that PLRN achieves competitive performance over existing methods with less computation time and memory.

\end{abstract}

\section{Introduction}

As the number of videos from Internet Protocol cameras (IP cameras) for video surveillance increases, understanding video contents such as action recognition~\cite{ChaudharyDPM19, kim2019skeleton, liu2019learning} and action localization~\cite{JargalsaikhanLO17, zeng2019graph} becomes crucial.
Moreover, videos with textual descriptions (\eg, titles, captions, or keywords) have encouraged research on multi-modal problems such as video captioning~\cite{krishna2017dense, sigurdsson2016hollywood} and temporal video grounding~\cite{chen2019semantic, gao2017tall, mun2020local, yuan2019find}.
Temporal video grounding is a challenging vision task, which aims to identify a grounding boundary semantically relevant to a sentence query in a long and untrimmed video.
\myfootnote{978-1-6654-3396-9/21/\$31.00 ©2021 IEEE}

\begin{figure}[t!]
  \centering
  \includegraphics[width=0.9\linewidth]{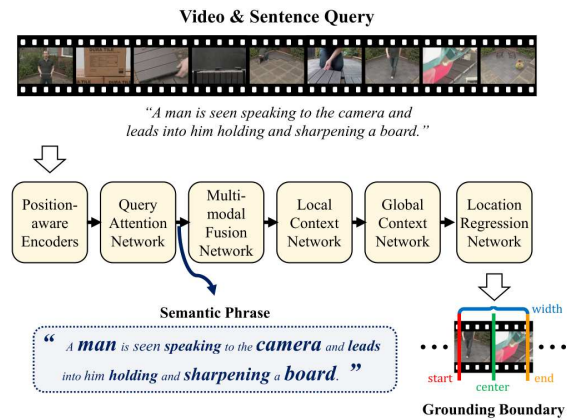}
  \caption{Temporal video grounding with a long and untrimmed video and a sentence query.
  Our network leverages a semantic phrase feature to understand semantic information of the query.
  After incorporating comprehensive contexts in the semantic phrase feature and encoded video, our network predicts a grounding boundary of the video semantically relevant to the query.}
\label{fig:temporal_video_grounding}
\end{figure}

Most existing methods~\cite{anne2017localizing, chen2019semantic,gao2017tall, wang2020temporally,xu2019multilevel,zhang2019man} are based on a propose-and-rank framework, which compares candidate proposals of a grounding boundary with a sentence query and selects one proposal having the highest matching score.
This approach based on the propose-and-rank framework only focuses on global representations of the query, which include some redundant word representations that are not helpful for grounding.
Recently, to tackle this issue, representations of a semantic phrase in the query have been used for temporal video grounding~\cite{mun2020local, yuan2019find}.
The semantic phrase is a sequence of words that are considered important clues to understand relationships between the sentence query and visual scenes (\ie, semantic entities such as actors, actions, objects, and backgrounds) ~\cite{hu2017modeling,yang2019dynamic,yu2018mattnet}.
However, these methods using a semantic phrase ignore contextual information for the semantic phrase~\cite{yuan2019find} or require heavy computation for extracting multiple semantic phrases~\cite{mun2020local}.

In contrast, we develop a network called Position-aware Location Regression Network (PLRN) by leveraging comprehensive contextual information for a single semantic phrase.
In PLRN, we introduce position-aware features from the query and video to understand positional information of words and video segments.
Also, we adopt a center-width location regression which predicts center and width values of a grounding boundary as well as start and end values of a grounding boundary for more efficient grounding.
Specifically, PLRN consists of multiple steps as depicted in Fig.~\ref{fig:temporal_video_grounding}.
First, we obtain position-aware features from a sentence query and a long and untrimmed video by using feature extraction and a position embedding strategy.
Then, leveraging the position-aware features, we make a sentence feature, word features, and video segment features.
We use the sentence feature and word features to generate a semantic phrase feature through an attention mechanism.
For multi-modal fusion, the semantic phrase feature and video segment features are merged into a multi-modal feature using Hadamard product~\cite{kim2016hadamard}.
After obtaining the multi-modal feature, we produce a context-aware feature by reflecting the comprehensive contexts which consist of 1) a local context and 2) a global context.
To consider the local context, we use a residual block~\cite{he2016deep} to make adjacent segment features of each segment in the multi-modal feature cover a wider field of view.
Also, to capture global representations with attention, we use non-local blocks~\cite{wang2018non} for the global context.
Finally, through the context-aware feature, we regress start, end, center, width values of a grounding boundary.
During training, we leverage a center-width location regression loss as well as a start-end location regression loss for efficient grounding.
We evaluate PLRN on two datasets: Charades-STA~\cite{gao2017tall} and ActivityNet Captions~\cite{krishna2017dense}.
PLRN can achieve competitive performance over existing methods with less computation time and memory because PLRN only exploits one semantic phrase rather than multiple semantic phrases.

\section{Related Work}

Existing methods for temporal video grounding can be largely grouped into two approaches: a proposal-based approach and a proposal-free approach.

\subsection{Proposal-based Video Grounding}
\label{sec:proposal_based_video_grounding}


A proposal-based approach~\cite{anne2017localizing,chen2019semantic,gao2017tall, wang2020temporally,xu2019multilevel,zhang2019man} is based on a propose-and-rank framework, where candidate proposals are obtained using sliding windows from a given video.
Then, the approach ranks the proposals according to the matching scores between the proposal and a sentence query.
Finally, the proposal with the highest score is selected.
Gao \etal~\cite{gao2017tall} and Hendricks \etal~\cite{anne2017localizing} use the sliding windows to make proposals and then compare each proposal with the query in a shared embedding space.
For better quality of the proposals, Xu \etal~\cite{xu2019multilevel} generate query-guided proposals by incorporating semantic information of the query into a proposal generation process.

The proposal-based approach is simple but has two limitations.
First, this approach has to compare all proposal-query pairs, which leads to heavy computational costs.
Second, the performance for temporal video grounding highly relies on quality of obtained proposals.
To avoid these limitations, our method is based on a proposal-free approach, where candidate proposals are not required.

\subsection{Proposal-free Video Grounding}
\label{sec:proposal_free_video_grounding}

A proposal-free approach~\cite{ghosh2019excl, liu2018temporal, mun2020local, rodriguez2020proposal, yuan2019find} directly locates grounding boundaries corresponding to the sentence query using regression.
Ghosh \etal~\cite{ghosh2019excl} focus on directly predicting the start and end values of a grounding boundary using regression.
For further improvement, Opazo \etal~\cite{rodriguez2020proposal} use a dynamic filter and model label uncertainty to transfer language information.
To capture semantic information in the query, Yuan \etal~\cite{yuan2019find} and  Mun \etal~\cite{mun2020local} propose to extract semantic phrase representations from the query.
Yuan \etal~\cite{yuan2019find} is based on a multi-modal co-attention mechanism using a semantic phrase.
Mun \etal~\cite{mun2020local} use multiple semantic phrases and multi-level video-text interaction for regression.
Also, a reinforcement learning (RL)-based approach~\cite{he2019read, wang2019language} is introduced for temporal video grounding, where the RL agent adjusts the predicted grounding boundary according to the learned policy.

Following the approach using semantic phrase representations~\cite{mun2020local,yuan2019find}, we leverage a semantic phrase.
However, the previous approach using a semantic phrase lacks comprehensive contextual information for the semantic phrase or requires heavy computational costs for multiple semantic phrases.
In contrast, we can capture comprehensive contextual information with only one semantic phrase through a position embedding strategy and a center-width location regression loss, which enables efficient temporal video grounding.

\section{Proposed Method}

\begin{figure*}[t!]
  \centering
  \includegraphics[width=0.94\linewidth]{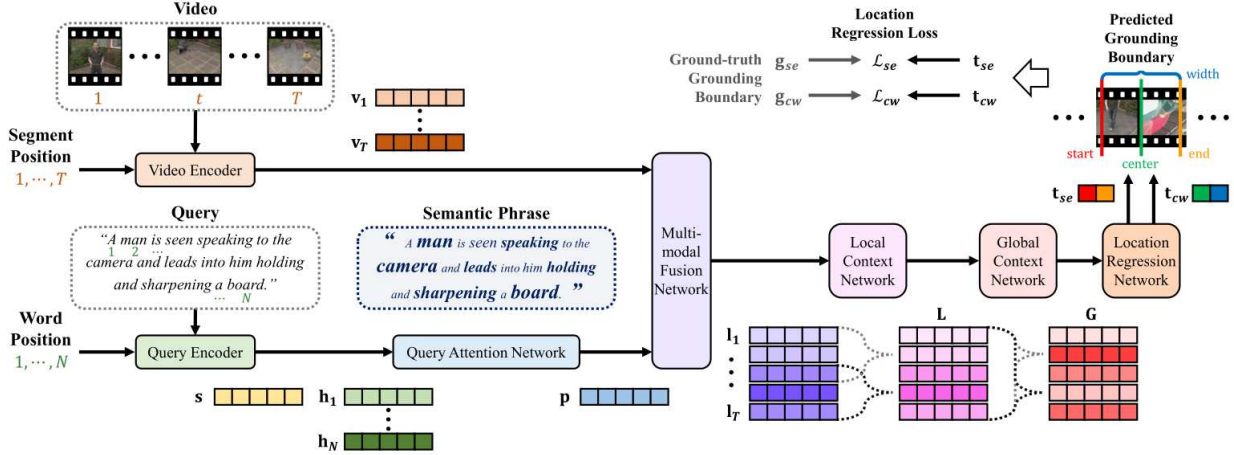}
  \caption{The overall architecture of the proposed network.}
\label{fig:framework}
\end{figure*}

When a long and untrimmed video and a sentence query are provided, the goal of temporal video grounding is to localize a grounding boundary of video segments described by a query.
Each video-query pair has one ground-truth, which is a grounding boundary ($g_{s},g_{e}$) starting at $g_{s}$ and ending at $g_{e}$.
The overall architecture of a proposed network, Position-aware Location Regression Network~(PLRN), is shown in Fig.~\ref{fig:framework}.
PLRN performs the following steps for temporal video grounding.
First, a sentence feature, word features, and video segment features are extracted via Position-aware Encoders (PE), which are represented as Video Encoder and Query Encoder.
In Video Encoder and Query Encoder, a position embedding strategy is used to incorporate positional information of video segments and words into the features, respectively.
Then, the semantic phrase feature is obtained through Query Attention Network (QAN) using an attention mechanism.
The semantic phrase feature and video segment features are combined into a multi-modal feature via Multi-modal Fusion Network (MFN).
The multi-modal feature is made into a context-aware feature through Local Context Network (LCN) and Global Context Network (GCN).
Finally, we predict a grounding boundary from the context-aware feature by Location Regression Network (LRN).

\subsection{Position-aware Encoders}
\label{sec:position_aware_encoders}

Position-aware Encoders receive a query and a video to make a sentence feature, word features, and video segment features. 
Then, the features are used as input to Query Attention Network.
In a sentence query, every word is made into a lowercase letter.
Given the sentence query with $N$ words, we use a trainable embedding matrix to extract query embedding matrix $\mathbf{Q}_{em}$ of which each column vector becomes an embedding of each word.
If there are more than 25 words in a sentence query, we truncate the sentence query to 25 words.
In addition, to capture positional information of the sentence query, we employ a position embedding strategy ~\cite{devlin2018bert}.
To this end, we encode the $n$-th word position into a position feature vector $f_{pos,q}(n)$ via a trainable mapping function $f_{pos,q}(\cdot)$. 
Then, each position feature vector becomes a column vector of a position feature matrix $\mathbf{P}_{q}=f_{pos,q}([1,2,\dots,N])$ which is added to the query embedding $\mathbf{Q}_{em}$ as
\begin{equation} 
\label{eq:query_position_embedding}
  \mathbf{Q} =\mathbf{Q}_{em}+\mathbf{P}_{q}\text{,}
\end{equation}
where $\mathbf{Q}$ is a position-aware query feature matrix.
The position-aware query feature $\mathbf{Q}$ is made into a sentence feature and word features using a bi-directional LSTM~(Bi-LSTM)~\cite{schuster1997bidirectional}.
Bi-LSTM yields hidden states given by
\begin{equation}
\label{eq:bi_LSTM}
  \{\,\mathbf{h}_n =\overrightarrow{\mathbf{h}_n}\,||\,\overleftarrow{\mathbf{h}_n}\,|\,n=1,2, \dots , N\,\} = \mathrm{Bi\text{-}LSTM}(\mathbf{Q}) \text{,}
\end{equation}
where $\mathbf{h}_n\in \mathbb{R}^{d}$ is a concatenation ($||$) of two hidden states in the $n$-th blocks with forward $(\overrightarrow{\mathbf{h}_n})$ and backward $(\overleftarrow{\mathbf{h}_n})$ direction.
The dimension $d$ of hidden states in LSTM is set to 512.
The $n$-th hidden state $\mathbf{h}_n$ becomes the $n$-th word feature.
The sentence feature $\mathbf{s}$ is obtained by a concatenation of the last hidden states of forward and backward direction:
\begin{equation}
\label{eq:sentence_feature}
  \mathbf{s} = \overrightarrow{\mathbf{h}_N}\,||\,\overleftarrow{\mathbf{h}_1} \in \mathbb{R}^{d} \text{.}
\end{equation}

Given a long and untrimmed video $\mathbf{V}_{raw}$, we divide the video $\mathbf{V}_{raw}$ into $T$ segments, where
each segment consists of a fixed length of frames and overlaps with adjacent segments by half the length of the segment.
Specifically, the length of each segment is 16 frames and the number of segments $T$ is set to 128.
If the number of segments is less than $T$ due to a short video, empty segments filled with zero values are added to the video segments.
Then, a video embedding $\mathbf{V}_{em}$ is extracted from video segments via 3D CNN-based feature extractors.
Depending on datasets, we use I3D~\cite{carreira2017quo} or C3D~\cite{tran2015learning} as 3D CNN-based feature extractors.
The extracted video embedding $\mathbf{V}_{em}$ can be written as
\begin{equation}
\label{eq:video_encoding}
  \mathbf{V}_{em} = \mathrm{ReLU}\left(\,\mathbf{W}_{em}\, f_{cnn}(\mathbf{V}_{raw})\,\right) \text{,}
\end{equation}
where $\mathbf{W}_{em}$ is trainable weights for reducing the feature dimension extracted by 3D CNN-based feature extractor $f_{cnn}(\cdot)$.
In addition, to capture positional information of the video, we employ a position embedding strategy~\cite{devlin2018bert}.
To this end, we encode the $t$-th segment position into a position feature vector $f_{pos,v}(t)$ via a trainable mapping function $f_{pos,v}(\cdot)$. 
Then, each position feature vector becomes a column vector of a position feature matrix $\mathbf{P}_{v}=f_{pos,v}([1,2,\dots,T])$ which is added to the video embedding $\mathbf{V}_{em}$ as
\begin{equation}  
\label{eq:video_position_embedding}
  \mathbf{V} =\mathbf{V}_{em}+\mathbf{P}_{v}  \text{,}
\end{equation}
where $\mathbf{V}$ is a position-aware video feature matrix.
Then, the position-aware video feature can be represented as $\mathbf{V}= [\,\mathbf{v}_1,\mathbf{v}_2, \dots, \mathbf{v}_T\,] \in \mathbb{R}^{d\times T} $, where $\mathbf{v}_t$ is the $t$-th video segment feature and $d$ is the dimension of the video segment feature.

\subsection{Query Attention Network}
\label{sec:query_attention_network}

Motivated by~\cite{mun2020local, yuan2019find}, we extract a semantic phrase feature from the query through an attention mechanism.
For the semantic phrase feature $\mathbf{p}$, we use the sentence feature $\mathbf{s}$ and the word features $\{\mathbf{h}_n\}_{n=1}^N$.
The semantic phrase feature $\mathbf{p} \in \mathbb{R}^{d}$ is given by
\begin{align}
  \label{eq:guidance_vector}
  &\mathbf{g} = \mathrm{ReLU}\left(\mathbf{W}_\mathbf{gs}\,\mathbf{s}\right) \in \mathbb{R}^d \text{,} \\
  \label{eq:query_attention_score}
  &\mathbf{\alpha}_n = \mathbf{w}_{qat}^\top ~\mathrm{tanh}\left(\,\mathbf{W}_\mathbf{\alpha g}\,\mathbf{g}+\mathbf{W}_\mathbf{\alpha h}\,\mathbf{h}_n\,\right) \text{,} \\
  \label{eq:query_attention_weight}
  &\mathbf{a} = \mathrm{softmax}\left([\,\mathbf{\alpha}_1, \mathbf{\alpha}_2,\dots, \mathbf{\alpha}_N\,]\right)^\top\in \mathbb{R}^N \text{,} \\
  \label{eq:semantic_phrase}
  &\mathbf{p} = \sum_{n=1}^N a_n \mathbf{h}_n \mathrm{,}\;
\end{align}
where $\mathbf{W}_\mathbf{yx}$ indicates a trainable weight matrix for transformation from $\mathbf{x}$ to $\mathbf{y}$, $\mathbf{w}_{qat}$ denotes a trainable weight vector for query attention, and the $n$-th element $a_n$ of an attention weight vector~$\mathbf{a}$ indicates
an attention weight for the $n$-th word.

\subsection{Multi-modal Fusion Network}
\label{sec:multi_modal_fusion_network}
For multi-modal fusion, The semantic phrase feature $\mathbf{p}$ and video segment features $\mathbf{V} = [\,\mathbf{v}_1,\mathbf{v}_2, \dots, \mathbf{v}_T\,]$ are merged. 
To this end, we use Hadamard product~\cite{kim2016hadamard} of the $t$-th video segment feature $\mathbf{v}_t $ and the semantic phrase feature $\mathbf{p}$ as
\begin{equation}
  \label{eq:modality_fusion}
  \mathbf{l}_{t}=\mathbf{W}_{mf}\left(\,\mathbf{W}_{\mathbf{lv}}\,\mathbf{v}_t\odot\mathbf{W}_{\mathbf{lp}}\,\mathbf{p}\,\right) \text{,}
\end{equation}
where $\mathbf{l}_{t} \in \mathbb{R}^{d}$ denotes a multi-modal feature vector, 
$\mathbf{W}_{mf}$ indicates a trainable weight matrix for multi-modal fusion, and $\odot$ is the Hadamard product operator.

\subsection{Local Context Network.}
\label{sec:local_context_network}
Each video segment lacks a field of view due to divided video segments with a fixed length of frames.
To incorporate a wider field of view of each segment, we aggregate adjacent segment features using a residual block (ResBlock)~\cite{he2016deep}.
Specifically, the residual block with large bandwidth~(\ie, 15) is used.
Then, a local context-aware matrix $\mathbf{L}\in \mathbb{R}^{d\times T}$ is produced by
\begin{equation}
\label{eq:res_block}
  \mathbf{L}=\mathrm{ResBlock}\big([\,\mathbf{l}_{1}, \mathbf{l}_{2}, \dots, \mathbf{l}_{T}\,]\big)
  \text{.}
\end{equation}

\subsection{Global Context Network.}
\label{sec:global_context_network}
To capture global representations with attention, we employ non-local blocks~\cite{wang2018non}.
Specifically, two non-local blocks with four heads are used.
Using the local context-aware matrix $\mathbf{L}$ as input to the non-local blocks, we produce a global context-aware feature matrix $\mathbf{G}\in \mathbb{R}^{d\times T}$, which is represented as
\begin{equation}
\label{eq:nl_block}
  \mathbf{G} = \mathbf{L}+\mathbf{W}_{val}\,\mathbf{L}\,\mathrm{softmax}\left(\,\frac{(\mathbf{W}_{qry}\mathbf{L})^\top(\mathbf{W}_{key}\mathbf{L})}{\sqrt{d}}\,\right)^\top
  \text{,}
\end{equation}   
where $\mathbf{W}_{val}$, $\mathbf{W}_{qry}$, and $\mathbf{W}_{key}$ are trainable weight matrices for value, query, and key in the non-local block.

\subsection{Location Regression Network}
\label{sec:location_regression_network}

Leveraging the global context-aware feature matrix $\mathbf{G}$, we make a semantics-aware feature $\mathbf{r}$ as 
\begin{align}
\label{eq:temporal_attentive_pooling}
  &\mathbf{b} = \mathrm{softmax}\left(\,\mathbf{w}_{tat}^\top ~\mathrm{tanh}(\mathbf{W}_{\mathbf{bG}}\,\mathbf{G})\,\right)^\top\in \mathbb{R}^T \text{,}  \\
  \label{eq:semantics_aware_feature}
  &\mathbf{r} = \mathbf{G}\mathbf{b}
  \text{,}
\end{align}
where an attention weight vector~$\mathbf{b}$ highlights important segments for grounding using temporal attentive pooling and $\mathbf{w}_{tat}$ denotes a trainable weight vector for the attentive pooling.
Finally, start and end values of a grounding boundary $\mathbf{t}_{se}=[\tau_s,\tau_e]^\top\in \mathbb{R}^2$ are predicted from a Multi-Layer Perceptron~(MLP) as 
\begin{align}
  \label{eq:regression}
  &\mathbf{t}_{se} =\mathrm{ReLU}\left(\,\mathbf{W}_{reg_{se}}\,\mathrm{ReLU}(\mathbf{W}_{\mathbf{t}_{se}\mathbf{r}}\,\mathbf{r})\,\right)
  \text{,}
\end{align}
where $\mathbf{W}_{reg_{se}}$ indicates a trainable weight matrix for the start-end location regression.
For more efficient grounding, we also predict center and width values of a grounding boundary $\mathbf{t}_{cw}=[\tau_c,\tau_w]^\top\in \mathbb{R}^2$:
\begin{equation}
\label{eq:cw_regression}
  \mathbf{t}_{cw} =\mathrm{ReLU}\left(\,\mathbf{W}_{reg_{cw}}\,\mathrm{ReLU}(\mathbf{W}_{\mathbf{t}_{cw}\mathbf{r}}\,\mathbf{r})\,\right)
  \text{,}
\end{equation}
where $\mathbf{W}_{reg_{cw}}$ indicates a trainable weight matrix for the center-width location regression.

\subsection{Training}
\label{sec:training}

For training, we use three losses: 1) a start-end location regression loss $\mathcal{L}_{se}$, 2) a center-width location regression loss $\mathcal{L}_{cw}$, and 3) a temporal attention calibration loss $\mathcal{L}_{tem}$.
The total losses $\mathcal{L}$ are represented as
\begin{equation}
\label{eq:loss}
  \mathcal{L} = \mathcal{L}_{se} + \mathcal{L}_{cw} + \mathcal{L}_{tem}
  \text{.}
\end{equation}
The location regression losses (\ie, $\mathcal{L}_{se}$ and $\mathcal{L}_{cw}$) help directly localizing a ground-truth grounding boundary and the temporal attention calibration loss $\mathcal{L}_{tem}$ promotes better temporal attention.

\begin{table*}
\centering
  \caption{Ablation study of different components and losses in the proposed network on Charades-STA.} 
  \label{tab:ablation_module}
  \begin{tabular}{L{0.22\textwidth} | C{0.1\textwidth} | C{0.05\textwidth}C{0.05\textwidth} |C{0.055\textwidth}C{0.055\textwidth}|cccc}
    \hline
    \multirow{2}{*}{Method} & \multirow{2}{*}{Query Info.} & \multicolumn{2}{c|}{Losses} &\multicolumn{2}{c|}{Contexts} & \multirow{2}{*}{R@0.3} & \multirow{2}{*}{R@0.5} & \multirow{2}{*}{R@0.7} &\multirow{2}{*}{mIoU} \\ \cline{3-6}
    &  & $\mathcal{L}_{tem}$ & $\mathcal{L}_{cw}$ & Local & Global &  &  &  &  \\
    \hline 
    Baseline & $\mathbf{s}$ & - & - & - & - & 56.34 & 41.24 & 19.33 & 35.20 \\
    PLRN w/o QAN & $\mathbf{s}$ & \cmark & \cmark & \cmark & \cmark & 69.65 & 57.07 & 36.16 & 49.38 \\
    PLRN w/o $\mathcal{L}_{tem}$ & $\mathbf{p}$ & - & \cmark & \cmark & \cmark & 62.61 & 48.20 & 27.07 & 43.11 \\
    PLRN w/o $\mathcal{L}_{cw}$ & $\mathbf{p}$ & \cmark & - & \cmark & \cmark & 72.20 & 58.41 & 35.32 & 50.52 \\
    PLRN w/o LCN & $\mathbf{p}$ & \cmark & \cmark & - & \cmark & 67.47 & 53.39 & 29.44 & 46.64 \\
    PLRN w/o GCN & $\mathbf{p}$ & \cmark & \cmark & \cmark & - & 66.40 & 51.37 & 27.23 & 45.17 \\
    \hline
    PLRN & $\mathbf{p}$ & \cmark & \cmark & \cmark & \cmark & \textbf{73.17} & \textbf{59.73} & \textbf{37.28} & \textbf{51.74} \\
    \hline
  \end{tabular}
\\ \centering{\cmark denotes the use of the loss or the context.}
\end{table*}

\textbf{Location regression loss.}
Following the regression based on ~\cite{yuan2019find}, $\mathcal{L}_{se}$ and $\mathcal{L}_{cw}$ are the sum of smooth $\mathbf{L}_1$ distances~\cite{girshick2015fast} between the normalized ground-truth grounding boundary and the predicted grounding boundary, which are defined as
\begin{align}
\label{eq:location_regression_loss}
  &\mathcal{L}_{se} = f_{sm}(g_s-\tau_s)+f_{sm}(g_e-\tau_e)
  \text{,} \\
  \label{eq:center_width_location_regression_loss}
  &\mathcal{L}_{cw} = f_{sm}(g_c-\tau_c)+f_{sm}(g_w-\tau_w)
  \text{,} \\
  \label{eq:smooth_L1_function}
  &f_{sm}(z) =\begin{cases}
    0.5z^2, & \text{if } |z|<1.\\
    |z|-0.5, & \text{otherwise}.
  \end{cases}
  \text{,}
\end{align}
where $f_{sm}(\cdot)$ is a smooth $\mathbf{L}_1$ function, $(g_s,g_e)$ is start and end values of the ground-truth grounding boundary, and $(g_c,g_w)$ is center and width values of the ground-truth grounding boundary.

\textbf{Temporal attention calibration loss.}
For better quality of learned temporal attention (\ref{eq:temporal_attentive_pooling}), we employ an attention calibration loss based on~\cite{yuan2019find}. 
The loss makes PLRN generate temporal attention relevant to the ground-truth grounding boundary:
\begin{align}
\label{eq:temporal_attention_calibration_loss}
  &\mathcal{L}_{tem} = -\frac{\sum_{t=1}^T \phi_t\,\mathrm{log}(b_t)}{\sum_{t=1}^T\phi_t}
  \text{,} \\
  \label{eq:temporal_attention_calibration_indicator}
  &\phi_t =\begin{cases}
    1, & \text{if the $t$-th segment is in ($g_{s},g_{e}$)}.\\
    0, & \text{otherwise}.
  \end{cases}
  \text{,}
\end{align}
where $b_t$ is the $t$-th element of $\mathbf{b}$ in the temporal attentive pooling (\ref{eq:temporal_attentive_pooling}).

\section{Experimental Results}

\subsection{Datasets}
\label{sec:datasets}

\subparagraph{\textbf{Charades-STA}}
is a widely used dataset for temporal video grounding. 
It has originally been collected from the Charades dataset~\cite{sigurdsson2016hollywood}, which is a benchmark for action recognition and video captioning.
Gao \etal~\cite{gao2017tall} gathered query annotations in the Charades dataset and then made the Charades-STA dataset for temporal video grounding.
The data splitting strategy in~\cite{gao2017tall} is adopted for our experiments.
There are 6,672 videos with 16,128 video-query pairs, which are composed of 12,408 and 3,720 pairs for training and testing, respectively.
Each video is about 29.76 seconds long and has 2.4 annotated moments on average.
The vocabulary size is 1,140.
For the Charades-STA, video segment features are extracted using a pre-trained 3D CNN-based feature extractor I3D~\cite{carreira2017quo}, where parameters are fixed during training.

\subparagraph{\textbf{ActivityNet Captions}}
is a widely used dataset for dense video captioning~\cite{krishna2017dense}.
Providing query annotations, the ActivityNet Captions can be used for temporal video grounding.
The dataset contains 20,000 videos that are related to 200 activity classes with 100,000 queries.
The video length is about 120 seconds. 
Each video contains 3.65 queries and ground-truth boundaries, and each query contains 13.48 words on average.
The vocabulary size is 11,125.
For the ActivityNet Captions, video segment features are extracted using a pre-trained 3D CNN-based feature extractor C3D~\cite{tran2015learning}, where parameters are fixed during training.
The original data splitting strategy divides the dataset into 10,024, 4,926, and 5,044 videos for training, validation, and testing, respectively.
There are various data splitting strategies because the testing set is not publicly available.
Following the strategy in ~\cite{mun2020local}, we test the proposed method on two combined validation sets ($val_1$ and $val_2$).
Also, other strategies that test the method on $val_1$ and $val_2$ separately are used to make fair comparisons with the other existing methods~\cite{he2019read, wu2020tree, zhang2019cross}.

\subsection{Experiment Setup}
\label{sec:experiment_setup}

For evaluation, we follow \cite{gao2017tall} that uses two metrics: 1) recall at three thresholds of the temporal Intersection over Union~(R@tIoU) and 2) mean of all tIoUs~(mIoU).
The R@tIoU is the percentage of testing data with the tIoU between a ground-truth boundary and a predicted boundary larger than the threshold.
For R@tIoU thresholds, we adopt $0.3$, $0.5$, and $0.7$.
The tIoU between the normalized ground-truth grounding boundary~($g_{s},g_{e}$) and the predicted one~($\tau_{s},\tau_{e}$) is defined as
\begin{equation}
\label{eq:temporal_iou}
  \mathrm{tIoU}=\frac{\mathrm{min} (g_{e},\tau_{e})-\mathrm{max}(g_{s},\tau_{s})}{\mathrm{max}(g_{e},\tau_{e})-\mathrm{min}(g_{s},\tau_{s})}
  \text{.}
\end{equation}
For training, we use Adam optimizer~\cite{kingma2014adam} and set the learning rate and mini-batch size to 0.0004 and 100, respectively.

\subsection{Ablation Study}
\label{sec:ablation_study}
\begin{table}[ht]
\centering
  \caption{Ablation study of different position embedding.} 
  \label{tab:ablation_position_embedding}
  \begin{tabular}{C{0.14\textwidth}C{0.12\textwidth}|cc}
    \hline
    \multicolumn{2}{c|}{Position Embedding Type} & \multirow{2}{*}{R@0.5} & \multirow{2}{*}{mIoU} \\ \cline{1-2}
    Video Segment & Word & &  \\
    \hline
     - & - & 48.63 & 44.13 \\ 
     \cmark & - & 56.45 & 48.91 \\ 
     - & \cmark & 55.03 & 48.50 \\ 
     \cmark & \cmark & \textbf{59.73} & \textbf{51.74}  \\ 
    \hline
  \end{tabular}
  \\ \centering{\cmark denotes the use of the position embedding}
\end{table}

\begin{table}[ht]
\centering
  \caption{Performance comparisons on Charades-STA.} 
  \label{tab:comparison_charades}
  \begin{tabular}{L{0.125\textwidth}|cccc}
    \hline
    Method & R@0.3 & R@0.5 & R@0.7 & mIoU\\
    \hline
    CTRL~\cite{gao2017tall} & - & 21.42 & 7.15 & -    \\ 
    SMRL~\cite{wang2019language} & - & 24.36 & 11.17 & -   \\ 
    SAP~\cite{chen2019semantic} & - & 27.42 & 13.36 & -   \\ 
    ACL~\cite{ge2019mac} & - & 30.48 & 12.20 & -   \\ 
    MLVI~\cite{xu2019multilevel} & 54.70 & 35.60 & 15.80 & -   \\ 
    TripNet~\cite{hahn2019tripping} & 51.33 & 36.61 & 14.50 & -   \\
    RWM~\cite{he2019read} & - & 36.70 & - & -   \\ 
    CBP~\cite{wang2020temporally} & 50.19 & 36.80 & 18.87 & 35.74   \\
    ExCL~\cite{ghosh2019excl} & 65.10 & 44.10 & 22.60 & -    \\
    TSP-PRL~\cite{wu2020tree} & - & 45.30 & 24.73 & 40.93   \\ 
    MAN~\cite{zhang2019man} & - & 46.53 & 22.72 & -   \\ 
    TMLGA~\cite{rodriguez2020proposal} & 67.53 & 52.02 & 33.74 & 48.22   \\
    SCDM~\cite{yuan2020semantic} & - & 54.44 & 33.43 & -   \\ 
    LGI~\cite{mun2020local} & \underline{72.96} & \underline{59.46} & \underline{35.48} & \underline{51.38}   \\ 
    \hline
    PLRN (Ours) & \textbf{73.17} & \textbf{59.73} & \textbf{37.28} & \textbf{51.74}   \\ 
    \hline
  \end{tabular}
\end{table}

\begin{table}[ht]
\centering
  \caption{Performance comparisons on ActivityNet.}
  \label{tab:comparison_activitynet}
  \begin{tabular}{L{0.125\textwidth}|cccc}
    \hline
    Method & R@0.3 & R@0.5 & R@0.7 & mIoU\\
    \hline
    MCN~\cite{anne2017localizing} & 21.37 & 9.58 & - & 15.83    \\ 
    CTRL~\cite{gao2017tall} & 28.70 & 14.00 & - & 20.54    \\ 
    ACRN~\cite{liu2018attentive} & 31.29 & 16.17 & - & 24.16    \\
    MLVI~\cite{xu2019multilevel} & 45.30 & 27.70 & 13.60 & -   \\ 
    TripNet~\cite{hahn2019tripping} & 48.42 & 32.19 & 13.93 & -   \\
    TMLGA~\cite{rodriguez2020proposal} & 51.28 & 33.04 & 19.26 & 37.78   \\
    CBP~\cite{wang2020temporally} & 54.30 & 35.76 & 17.80 & 36.85   \\
    ABLR~\cite{yuan2019find} & 55.67 & 36.79 & - & 36.99    \\
    LGI~\cite{mun2020local} & \underline{58.52} & \textbf{41.51} & \underline{23.07} & \underline{41.13}   \\ 
    \hline
    PLRN (Ours) & \textbf{58.60} & \underline{41.42} & \textbf{23.54} & \textbf{41.90}  \\
    \hline
    RWM~\cite{he2019read}$^\dagger$ & - & 36.90 & - & -   \\ 
    TSP-PRL~\cite{wu2020tree}$^\dagger$ & \underline{56.08} & \underline{38.76} & - & \underline{39.21}   \\ 
    \hline
    PLRN$^\dagger$ (Ours) & \textbf{57.25} & \textbf{40.63} & \textbf{23.12} & \textbf{40.27}  \\
    \hline
    CMIN~\cite{zhang2019cross}$^\ddagger$ & \underline{63.61} & \underline{43.40} & \underline{23.88} & -    \\
    \hline
    PLRN$^\ddagger$ (Ours) & \textbf{63.79} & \textbf{44.48} & \textbf{26.81} & \textbf{44.15}  \\
    \hline
  \end{tabular}
  \\ \centering{$^\dagger$: a $val_1$ set for testing, $~~$ $^\ddagger$: a $val_2$ set for testing}
\end{table}
\begin{table}[ht]
\centering
  \caption{Comparison of model size.} 
  \label{tab:comparison_model_size}
  \begin{tabular}{L{0.125\textwidth}|ccc}
    \hline
    Method & \# params & memory & time\\
    \hline
    LGI~\cite{mun2020local} & 33,818,355 & 129.01MB & 0.0623s\\ 
    \hline
    PLRN (Ours) & 15,920,373 & 60.73MB & 0.0271s \\
    \hline
  \end{tabular}
\end{table}

In this ablation study, in-depth analysis is provided to verify the effectiveness of different components and losses in PLRN.
In Table~\ref{tab:ablation_module}, the impact of components and losses in PLRN is investigated with the following variants:
1) Baseline,
2) PLRN w/o QAN,
3) PLRN w/o $\mathcal{L}_{tem}$,
4) PLRN w/o $\mathcal{L}_{cw}$,
5) PLRN w/o LCN, and
6) PLRN w/o GCN.
The Baseline is the network which uses the sentence feature $\mathbf{s}$ (\ref{eq:sentence_feature}) for multi-modal fusion rather than the semantic phrase feature $\mathbf{p}$ (\ref{eq:semantic_phrase}) and the multi-modal feature matrix $[\,\mathbf{l}_{1}, \mathbf{l}_{2}, \dots, \mathbf{l}_{T}\,]$ (\ref{eq:modality_fusion}) for location regression with only the start-end location regression loss.
PLRN w/o QAN is PLRN which replaces the semantic phrase feature $\mathbf{p}$ in Query Attention Network~(QAN) with the sentence feature $\mathbf{s}$ for multi-modal fusion.
PLRN w/o $\mathcal{L}_{tem}$ is PLRN without the temporal attention calibration loss $\mathcal{L}_{tem}$ (\ref{eq:temporal_attention_calibration_loss}).
PLRN w/o $\mathcal{L}_{cw}$ is PLRN without the center-width location regression loss $\mathcal{L}_{cw}$ (\ref{eq:center_width_location_regression_loss}).
PLRN w/o LCN is PLRN without Local Context Network~(LCN) using the residual block (\ref{eq:res_block}).
PLRN w/o GCN is PLRN without Global Context Network~(GCN) using the non-local blocks (\ref{eq:nl_block}).

From Table~\ref{tab:ablation_module}, the following results are made.
First, using the semantic phrase feature from the query attention network can improve performance by focusing on important words in the query.
Second, the temporal attention calibration loss contributes significantly to the performance by encouraging the network to focus on video segments within the ground-truth boundary. 
Third, the center-width location regression loss is useful for more efficient grounding.
Fourth, local and global contexts need to be captured for precise grounding because PLRN w/o LCN or GCN makes lower performance than PLRN.
In Table~\ref{tab:ablation_position_embedding}, we also analyze the impact of different position embedding types in PLRN on the Charades-STA.
Using the position embedding of both video segments (\ref{eq:video_position_embedding}) and words (\ref{eq:query_position_embedding}) in PLRN makes the best performance.
Excluding one of the position embedding types in PLRN degrades performance.
Therefore, we verify that positional information of video segments and words is meaningful and required for efficient grounding.

\begin{figure*}[t!]
  \centering
  \includegraphics[width=0.65\linewidth]{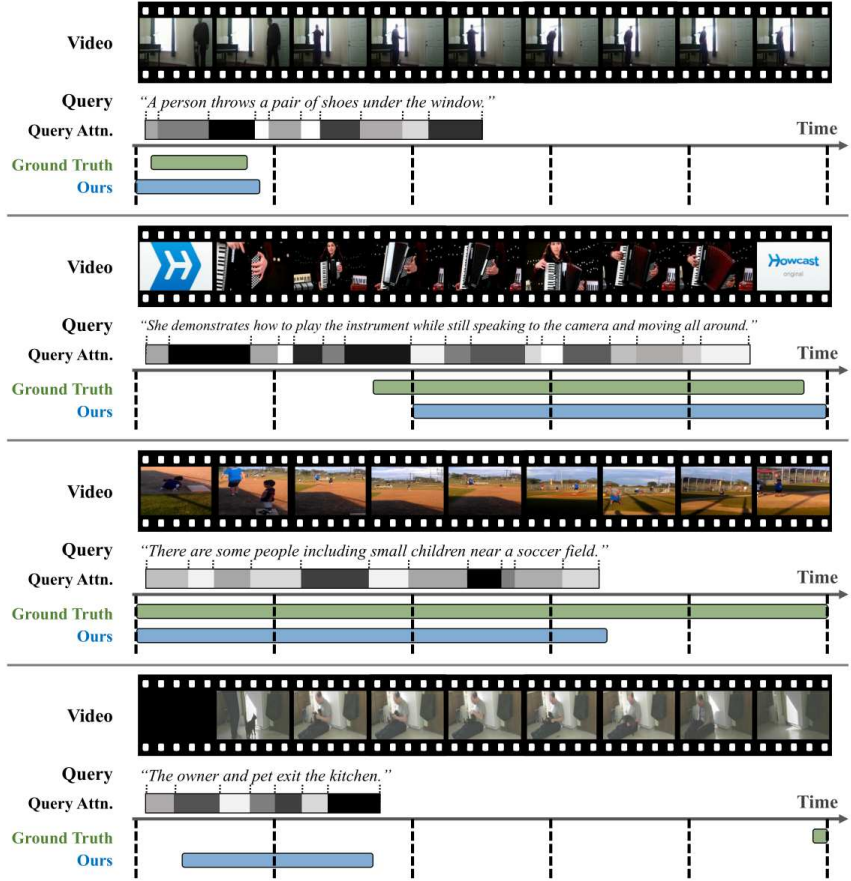}
  \caption{Qualitative results of PLRN on both Charades-STA and ActivityNet Captions datasets. 
  Given a pair of a video and a sentence query, PLRN yields a predicted grounding boundary (blue).
  We also visualize a ground-truth boundary (green) and query attention.
  In the query attention, a darker box means a higher attention value of the word in the query.
    }
\label{fig:qualitative_result}
\end{figure*}

\subsection{Comparison with Existing Methods}
\label{sec:comparison_with_existing_methods}

As observed in Tables~\ref{tab:comparison_charades}, our PLRN outperforms the existing methods on the Charades-STA.
PLRN surpasses LGI~\cite{mun2020local} by $1.8\%$ and $0.36\%$ in terms of R@0.7 and mIoU on the Charades-STA, respectively.
For the ActivityNet Captions dataset, because the testing set is not publicly available, we use three data splitting strategies:
1) testing on both validation sets of $val_1$ and $val_2$, 2) testing on $val_1$, which is denoted by $^\dagger$, and 3) testing on $val_2$, which is denoted by $^\ddagger$. 
In Table~\ref{tab:comparison_activitynet}, PLRN surpasses LGI by $0.47\%$ and $0.87\%$ in terms of R@0.7 and mIoU on the ActivityNet Captions, respectively.
Also, PLRN surpasses TSP-PRL~\cite{wu2020tree} and CMIN~\cite{zhang2019cross} by $1.87\%$ and $1.08\%$ in terms of R@0.5 on the separate $val_1$ and $val_2$ set of the ActivityNet Captions, respectively.
PLRN makes lower performance than LGI in R@0.5 on the ActivityNet Captions.
However, as observed in Table~\ref{tab:comparison_model_size}, PLRN can achieve competitive results with less memory and less computation time for forward propagation because PLRN uses only one semantic phrase rather than multiple semantic phrases.

\subsection{Qualitative Results}
\label{sec:qualitative_results}

Qualitative results are shown in Fig.~\ref{fig:qualitative_result}.
In the figure, we visualize ground-truth boundaries, predicted boundaries, and query attention $\mathbf{a}$ (\ref{eq:query_attention_weight}) of PLRN.
In the first and second examples of the figure, PLRN produces precise grounding boundaries with regard to ground-truth boundaries.
Also, PLRN performs an effective attention mechanism capturing semantic phrases which focus on `throws', `shoes', `window', `demonstrates', `play', and `instrument'.
In the third and fourth examples of the figure, PLRN fails to predict precise grounding boundaries with regard to ground-truth boundaries.
In the third example, the semantic phrase focuses on `including' and `near' which are not important words for temporal video grounding.
For the proper semantic phrase, `people' and `field' need to be captured with higher attention.
In the fourth example, the semantic phrase seems to be captured well.
However, PLRN fails to localize the ground-truth boundary because PLRN misunderstands that the owner and pet exit at the beginning of the video.

\section{Conclusions}
This paper addresses the problem of temporal video grounding, leveraging a semantic phrase from a sentence query.
We effectively aggregate information of the semantic phrase and video segments via local and global context modeling.
Also, we introduce position-aware encoders and a center-width location regression loss for more efficient grounding.
Therefore, the proposed network can capture comprehensive contextual information with only one semantic phrase, learning semantic representation of the query in an end-to-end manner.
We verify the effectiveness of the proposed method through ablation studies and comparisons with existing methods on both Charades-STA and ActivityNet Captions datasets.

\section*{Acknowledgement}
This work was supported by Institute of Information \& Communications Technology Planning \& Evaluation~(IITP) grant funded by the Korea government~(MSIT)~(No.2014-3-00123, Development of High Performance Visual BigData Discovery Platform for Large-Scale Realtime Data Analysis).

{\small
\bibliographystyle{ieee}
\bibliography{egbib}
}


\end{document}